\documentclass[11pt]{article}

\usepackage[margin=1in]{geometry}
\usepackage[T1]{fontenc}
\usepackage[utf8]{inputenc}
\IfFileExists{newpxtext.sty}
  {\usepackage{newpxtext,newpxmath}}
  {\usepackage{mathpazo}}
\usepackage{microtype}
\usepackage{graphicx}
\usepackage{booktabs}
\usepackage{multirow}
\usepackage[table]{xcolor}
\usepackage{amsmath}
\usepackage[most]{tcolorbox}
\usepackage{caption}
\usepackage{fancyhdr}
\usepackage[numbers,sort&compress]{natbib}
\usepackage{enumitem}
\usepackage[colorlinks=true]{hyperref}
\usepackage{orcidlink}

\definecolor{accent}{RGB}{63,110,158}     
\definecolor{takebg}{RGB}{241,245,249}    
\definecolor{hlrow}{RGB}{247,218,215}     
\definecolor{hlrowblue}{RGB}{224,235,244} 
\hypersetup{linkcolor=accent, citecolor=accent, urlcolor=accent}

\newtcolorbox{takeaway}[1]{enhanced, colback=takebg, frame hidden,
  borderline west={2.5pt}{0pt}{accent}, sharp corners,
  left=8pt, right=8pt, top=6pt, bottom=6pt,
  before upper={\textbf{#1.}~\itshape}}

\newcommand{\lead}[1]{\smallskip\noindent\textbf{#1.}~}

\title{\bfseries Temperon: Full-Time SAM Quality at a Third Less Wall-Clock}
\author{Stamatis Mastromichalakis\\
  \small Independent Researcher, tmnetworks.gr\\
  \small\texttt{stamatis@tmnetworks.gr}\\
  \small\orcidlink{0000-0003-1713-5078}~\url{https://orcid.org/0000-0003-1713-5078}}
\date{}

\begin{document}
\maketitle
\thispagestyle{fancy}

\begin{abstract}
\noindent
Sharpness-aware minimization (SAM) doubles the cost of every training step,
yet its benefit concentrates where training ends. We study \emph{where} an
expensive training mode should be spent and propose \textbf{Temperon}: a
plain-SGD explorer for the first 43\% of the epoch budget, then one
scheduled hand-off that gives the entire final cosine anneal to a
SAM-wrapped Muon refiner. On CIFAR-10/100, SVHN and Tiny ImageNet (five
seeds, times reported as epochs-to-target $\times$ an idle-GPU-calibrated
epoch cost), Temperon matches the best full-time-SAM recipe on accuracy
everywhere while reaching the hardest common target 35\%, 34\% and 32\%
sooner on three of the four, and sits a tier above the published SAM+SGD
recipe at level cost. Ablations make the attribution exact: the Muon
refiner is worth +0.85pp with everything else fixed; the explorer's shape
and its restarts are worth nothing, and we withdraw them as contributions.
Re-running the closest rival, late-phase SAM, at matched budget shows the
frontier: it is fastest to every mid-level target, but the tier the Muon
refiner buys (0.83 on CIFAR-100, 0.97 on CIFAR-10) is reached by no
SGD-refined method in any seed, and on Tiny ImageNet, where Muon buys no
tier, the rival simply wins --- the measured boundary of the method. The
allocation law transfers to GPT-2 pretraining (full-SAM quality at $-29\%$
wall-clock) and GLUE fine-tuning (never worse than full-time SAM at a third
of its SAM cost). Two constants organize the economics: skipping SAM early
buys a fixed credit, and a Muon epoch costs $1.50\times$ a SAM+SGD epoch on
all four datasets. Finally, the hand-off cannot be timed from the
trajectory: under cosine schedules the accuracy curve is plateau-then-surge,
so the information lives in the schedule, making the scheduled switch
principled rather than convenient. Code and a pip-installable
implementation are released.
\end{abstract}

\begin{center}
\emph{Official Repository:}
\href{https://github.com/MStamatis/temperon}{\emph{GitHub Repository}}
\end{center}

\section{Introduction}\label{sec:intro}

Two of the most effective additions to a modern training run are also two of
the most expensive. Sharpness-aware minimization \citep{foret2021sam}
doubles the cost of every step it touches by taking an ascent step before
each descent step; Muon \citep{jordan2024muon,liu2025muon}, applied to a
convolutional backbone, costs $2.4\times$ an SGD epoch in our measurements.
The standard way to use either is full-time, paying the surcharge from the
first epoch to the last. Yet the benefit is not spread over the run:
\citet{zhou2025latephase} show that a few epochs of SAM at the end of
training recover nearly all of full-time SAM's generalization, and our own
measurements show full-time SAM+Muon spending its middle fifty epochs on an
accuracy plateau it could have reached far more cheaply
(Figure~\ref{fig:plateau}).

This paper treats the question as one of \emph{allocation}: given a fixed
epoch budget and an expensive mode, where should the budget carry the
surcharge? The answer we converge on is two rules. \emph{Pay the expensive
mode only in the tail}, and \emph{let the tail own the entire final anneal}:
one scheduled hand-off at the boundary where the last learning-rate descent
begins, never inside it. We call the resulting recipe \textbf{Temperon},
after tempering, the controlled heat treatment applied after quenching that
sets a metal's final toughness, because that is what the expensive tail
governs: what happens during the last anneal.

We evaluate on four vision datasets with every baseline re-run inside one
pipeline (five seeds, calibrated timing), pretrain a GPT-2-class model,
fine-tune RoBERTa on four GLUE tasks, and re-run the closest published
rival at matched budget, including on the dataset where it beats us.

\lead{Contributions}
(i)~a scheduled hand-off recipe reaching full-time-SAM quality at
roughly a third less wall-clock on three of four vision datasets;
(ii)~exact attribution via two single-change ablations: the refiner sets the
accuracy tier (+0.85pp), the allocation shape and restarts do not;
(iii)~a measured frontier against the strongest published rival, including
the dataset where the rival wins;
(iv)~transfer of the allocation law to GPT-2 pretraining and GLUE
fine-tuning;
(v)~a cost model (credit against a flat $1.50\times$ premium) that
predicts which side of the trade a practitioner is on before training;
(vi)~evidence that the switch cannot be timed adaptively from the
trajectory, making the schedule the principled choice.

\begin{takeaway}{Paper in one line}
Pay for sharpness only in the tail, let the tail own the entire anneal, and
hand it to the strongest refiner the task rewards; everything else about
the schedule is negotiable, and we measured that it is.
\end{takeaway}

\section{Method}\label{sec:method}

\lead{The recipe}
One training run, two stages, one scheduled switch. The reference vision
configuration: (1)~an \emph{explorer}, plain Nesterov SGD (lr 0.1,
weight decay $5\times10^{-4}$), for the first 43\% of the epoch budget;
(2)~one scheduled \emph{hand-off} at that boundary: the incoming optimizer
inherits the momentum buffer, takes a 200-step learning-rate warmup, and
SAM's perturbation radius ramps from zero over 400 steps, so the switch
never shocks the loss; (3)~a \emph{refiner}, Muon (lr 0.01, weight decay
0.2) wrapped in SAM's two-pass ascent--descent ($\rho=0.05$), owning a
fresh cosine anneal for the remaining 57\%. Nothing in stage~1 turned out to
be load-bearing: we originally used a cyclic schedule with warm restarts
\citep{loshchilov2017sgdr}, and \S\ref{sec:attribution} shows a single
cosine is indistinguishable.

\lead{The tail is per-task}
The tail's base optimizer is the strongest refiner the task rewards: Muon on
CIFAR-10/100 and SVHN; SGD on Tiny ImageNet, where Muon buys no accuracy
tier (\S\ref{sec:rival}); Muon under a warmup--stable--decay (WSD) schedule
for GPT-2 pretraining, where the only expensive mode is SAM itself and the
hand-off is the SAM toggle at the decay boundary; AdamW for GLUE
fine-tuning. What is constant across domains is the two rules, not the
optimizers playing the roles.

\lead{The switch point is a dial, not a constant}
Moving the vision hand-off from epoch 43 to epoch 80 lands at
$0.8249\pm0.0022$ ($-0.42$pp against full SAM+Muon, $p{=}0.015$; level with
the published SAM+SGD recipe) for 43\% less total compute. A longer
tail buys the expensive method's ceiling at its cost; a shorter one trades a
measured slice of the ceiling for a much cheaper run. The LM and GLUE runs
use a 30\% tail.

\lead{The exchange rate}
Skipping SAM before the switch buys a fixed credit,
$43\times(c_{\text{SAM}}-c_{\text{cheap}})$; handing the tail to Muon pays a
premium, $(\text{tail epochs to target})\times
(c_{\text{SAM+Muon}}-c_{\text{SAM+SGD}})$. Measured on all
four datasets the ratio $c_{\text{SAM+Muon}}/c_{\text{SAM+SGD}}$ is
$1.504/1.508/1.495/1.497$, a constant we did not expect to be that flat,
so the premium nearly cancels the credit and the net cost against
full-time SAM+SGD lands at $-2.0\%/-2.1\%/-2.4\%$ with the accuracy
returned instead (+0.63/+0.27/+0.09pp). Tiny ImageNet keeps an SGD tail
(premium zero by construction) and banks the credit as $-32.6\%$ wall-clock.

\begin{figure}[t]
  \centering
  \includegraphics[width=0.72\linewidth]{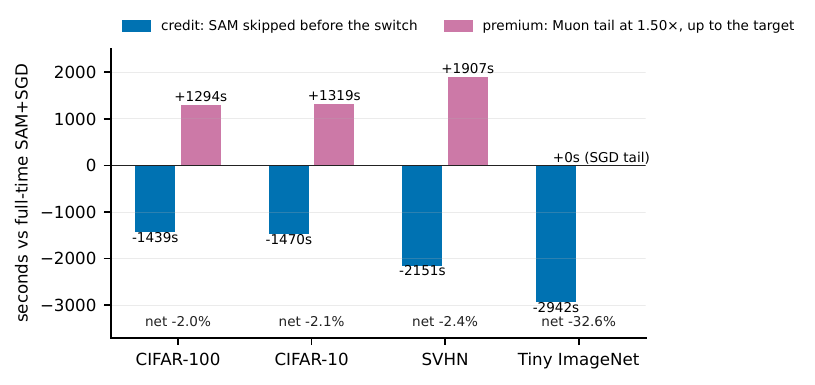}
  \caption{\textbf{The exchange rate.} The credit bought by skipping SAM
  before the switch (blue, negative) against the premium paid by the Muon
  tail (pink, positive), per dataset, at idle-GPU calibrated epoch costs.
  Three datasets pay the same $1.50\times$ premium out of the same credit
  and land at $-2\%$ net; Tiny ImageNet's SGD tail pays no premium and keeps
  the credit as time.}
  \label{fig:exchange}
\end{figure}

\section{Experimental setup}\label{sec:setup}

\lead{Vision}
CIFAR-10/100, SVHN and Tiny ImageNet, on a wide ResNet-110: the depth-110
topology of \citet{he2016resnet} at $4\times$ width (stages 64/128/256,
27.6M parameters against the standard 1.7M; no external
``ResNet-110'' number is comparable to ours), projection shortcuts,
batch-norm momentum 0.01, and a stride-2 stem only on Tiny ImageNet's
$64\times64$ inputs. 100 epochs, batch 128, bf16, gradient clip 2.0,
flip/cutout/colour-jitter augmentation. Every arm runs the same five seeds
(42, 1181241943, 958682846, 271828, 314159), and every baseline is re-run
inside this pipeline: the claims are about wall-clock, and wall-clock does
not survive transplantation across papers, implementations and hardware.

\lead{How times are measured}
Not with a stopwatch. Raw wall-clock on a workstation is not a property of
the method: a GPU that also drives a monitor loses 20--28\% of every epoch
to the desktop, and we measured the same optimizer costing anywhere from
14.3 to 21.9\,s/epoch depending only on when a grid happened to be
scheduled. Every time in this paper is therefore \emph{epochs-to-target}
(measured from the runs and immune to load) $\times$ \emph{calibrated
seconds-per-epoch} (measured separately on an idle GPU, per optimizer per
dataset). Raw wall-clock is kept alongside in the released logs.

\lead{Language model}
GPT-2 124M (12 layers, width 768) \citep{radford2019gpt2}, context 1024, on
WikiText-103 \citep{merity2016wikitext} with a 400M-token budget under a
WSD schedule (decay over the last 30\%); Muon (lr 0.02) on the matrix
parameters, AdamW (lr $6\times10^{-4}$) on the rest. One seed, with a
measured noise floor (\S\ref{sec:lm}).

\lead{Fine-tuning}
RoBERTa-base \citep{liu2019roberta} on RTE, MRPC, STS-B and CoLA
\citep{wang2018glue}; standard hyper-parameters (lr $2\times10^{-5}$, batch
16), five seeds, and the primary metric, best dev score, was fixed
before any result was seen. All experiments run in a Docker container on a
single NVIDIA RTX 5090.

\section{Vision results}\label{sec:vision}

\begin{table}[t]
  \centering\small
  \caption{\textbf{Headline vision results.} Final test accuracy (5-seed
  mean~$\pm$~sd) and calibrated time to the hardest target any method
  reaches, with the fraction of seeds reaching it. Bold marks the best
  accuracy and time per row. The highlighted row is the dataset the rival
  wins outright (\S\ref{sec:rival}).}
  \label{tab:headline}
  \begin{tabular}{lcccc}
    \toprule
    dataset (target) & Temperon \emph{(ours)} & late-phase SAM & SAM+SGD & SAM+Muon \\
    & & \emph{(rival, re-run)} & \emph{(published)} & \emph{(ours, full-time)} \\
    \midrule
    CIFAR-100 (0.82) & \textbf{0.8295$\pm$0.0034} & 0.8234$\pm$0.0011 & 0.8232$\pm$0.0031 & 0.8292$\pm$0.0021 \\
      & 4643s $\cdot$ 5/5 & \textbf{3022s} $\cdot$ 5/5 & 4679s $\cdot$ 4/5 & 7190s $\cdot$ 5/5 \\
    \rowcolor{hlrow}
    Tiny ImageNet (0.69) & 0.7003$\pm$0.0033 & 0.7020$\pm$0.0028 & \textbf{0.7027$\pm$0.0032} & 0.6838$\pm$0.0019 \\
    \rowcolor{hlrow}
      & 6359s $\cdot$ 5/5 & \textbf{5750s} $\cdot$ 5/5 & 9333s $\cdot$ 5/5 & never \\
    CIFAR-10 (0.968) & \textbf{0.9695$\pm$0.0008} & 0.9688$\pm$0.0006 & 0.9668$\pm$0.0005 & 0.9694$\pm$0.0008 \\
      & 4849s $\cdot$ 5/5 & \textbf{3121s} $\cdot$ 4/5 & 4990s $\cdot$ 1/5 & 7371s $\cdot$ 5/5 \\
    SVHN (0.98) & \textbf{0.9807$\pm$0.0005} & 0.9798$\pm$0.0005 & 0.9798$\pm$0.0004 & 0.9804$\pm$0.0003 \\
      & 6948s $\cdot$ 5/5 & \textbf{5003s} $\cdot$ 3/5 & 7029s $\cdot$ 3/5 & 5084s $\cdot$ 5/5 \\
    \bottomrule
  \end{tabular}
\end{table}

\begin{figure}[t]
  \centering
  \includegraphics[width=\linewidth]{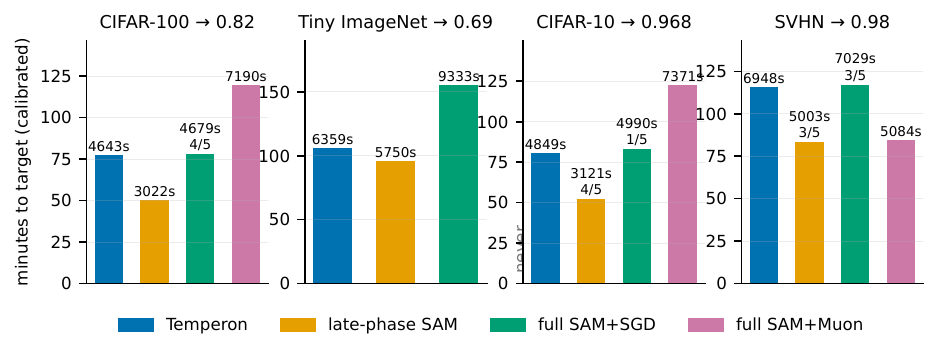}
  \caption{\textbf{Calibrated time to the headline target.} Four methods on
  four datasets; annotations give seconds and, where a method misses the
  target in some seeds, the fraction that reach it. The rival is fastest to
  every mid target; \S\ref{sec:rival} shows what it cannot reach.}
  \label{fig:frontier}
\end{figure}

\begin{figure}[t]
  \centering
  \includegraphics[width=\linewidth]{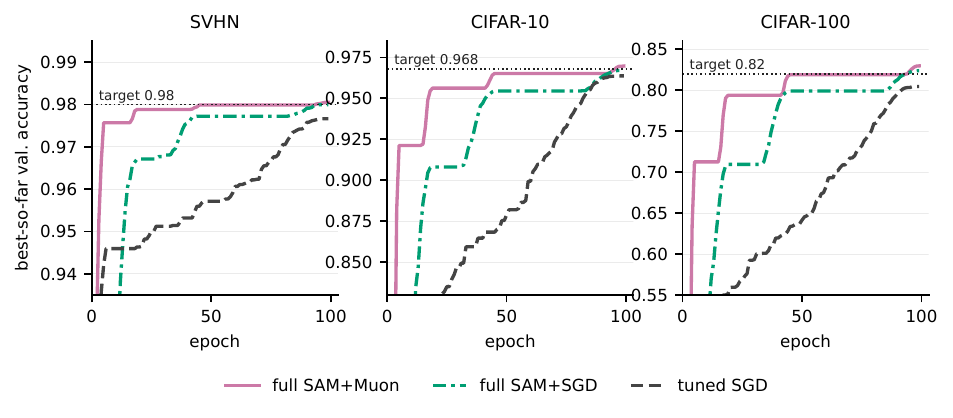}
  \caption{\textbf{Plateau-then-surge, and the one saturated task.}
  Mean best-so-far validation accuracy of the full-time arms against the
  headline bar (dotted). Full-time SAM+Muon plateaus by mid-run everywhere
  (its remaining progress arrives with the anneal), and only on SVHN
  does its plateau already clear the bar, which is why SVHN is the one
  dataset where it wins on time (Table~\ref{tab:headline}).}
  \label{fig:plateau}
\end{figure}

Table~\ref{tab:headline} and Figure~\ref{fig:frontier} carry the headline.
Temperon ties the best full-time-SAM recipe on final accuracy on all four
datasets (Welch $p=0.87/0.29/0.94/0.28$) while reaching the hardest common
target $35\%$ sooner on CIFAR-100, $34\%$ on CIFAR-10 and $32\%$ on Tiny
ImageNet. On SVHN it does not win on time, and Figure~\ref{fig:plateau}
shows why: SVHN's bar sits below full-time SAM+Muon's mid-run plateau, so
the expensive method crosses it by epoch 45 and never needs its anneal:
the one saturated case. Against the published SAM+SGD recipe specifically,
Temperon is ahead on accuracy on three of four datasets (+0.63pp
$p{=}0.016$, +0.27pp $p{<}0.001$, +0.09pp $p{=}0.015$) at level cost,
within 3\% on all four; that is the exchange rate of \S\ref{sec:method} at
work, not a coincidence. Allocation also dominates amount: uniform periodic
SAM at an equal-or-greater budget reaches only $0.8183\pm0.0016$ on
CIFAR-100, 1.12pp below the hand-off ($p{=}0.001$), while spending more
time.

\begin{takeaway}{Vision takeaway}
Temperon ties the best full-time-SAM recipe on accuracy on all four
datasets and reaches the hardest target $\sim$a third sooner on three; the
boundary cases (SVHN's saturated bar, Tiny ImageNet's failed Muon tier) are
measured and mechanistic, not anecdotal.
\end{takeaway}

\section{What earns the accuracy: attribution}\label{sec:attribution}

\begin{figure}[t]
  \centering
  \includegraphics[width=0.72\linewidth]{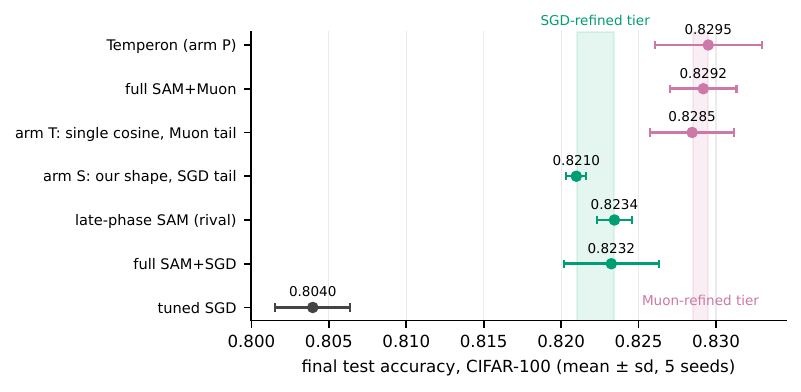}
  \caption{\textbf{The refiner sets the tier and nothing else does.}
  CIFAR-100 finals of every arm; shaded bands are the SGD-refined
  (0.8210--0.8234) and Muon-refined (0.8285--0.8295) tiers. Swapping only
  the refiner moves an arm between bands (+0.85pp, $p{=}0.005$); changing
  the explorer's shape or removing its restarts does not move it at all.}
  \label{fig:attribution}
\end{figure}

Temperon differs from late-phase SAM in two ways at once --- a Muon refiner,
and a cyclic explorer handing over to a fresh anneal --- so two ablations
swap one piece each while holding everything else fixed. \emph{Arm~S} puts
an SGD refiner behind our explorer: $0.8210\pm0.0006$, i.e.\ $-0.85$pp
($p{=}0.005$); \textbf{the refiner is the accuracy contribution}. The
same arm is 0.25pp \emph{worse} than the rival's plain mid-schedule switch
($p{=}0.005$) and needs seven more epochs to reach 0.82: the allocation
shape is \emph{not} a contribution, and earlier drafts of this work that
claimed it were wrong. \emph{Arm~T} keeps the Muon refiner and flattens the
explorer's four warm restarts into a single cosine: $0.8285\pm0.0027$,
$+0.10$pp for cyclic ($p{=}0.611$), identical epochs-to-target (0.80 at
epoch 91 and 0.82 at 94, both arms), and a statistical tie with full-time
SAM+Muon ($-0.07$pp, $p{=}0.655$): the restarts are an implementation
choice, not a mechanism. Figure~\ref{fig:attribution} shows the resulting
structure: every SGD-refined arm lands in 0.8210--0.8234, every Muon-refined
arm in 0.8285--0.8295, and nothing except the refiner moves an arm between
bands.

\begin{takeaway}{Attribution takeaway}
The Muon refiner is the contribution; the allocation shape is 0.25pp
\emph{worse} than a plain mid-schedule switch under an SGD refiner, and the
cyclic restarts are indistinguishable from a single cosine ($p{=}0.61$).
Earlier drafts claimed the shape; the claim was wrong and is withdrawn.
\end{takeaway}

\section{The closest rival, measured}\label{sec:rival}

\citet{zhou2025latephase} showed that SAM applied only late can match
full-time SAM. We re-run it at our tuned hyper-parameters and a matched SAM
budget (57 of 100 epochs) on all four datasets. It does two things at once:
one for our thesis and one against us.

\lead{For the thesis}
It reproduces, or outright beats, full-time SAM+SGD everywhere: within noise
on CIFAR-100 ($p{=}0.896$), Tiny ImageNet ($p{=}0.720$) and SVHN
($p{=}0.846$), and clearly above it on CIFAR-10 ($+0.20$pp, $p{<}0.001$,
against a published recipe that reaches the 0.968 target there in one seed
of five). An independent method, in our pipeline, arriving at the
allocation law from the other direction.

\lead{Against us}
It is the fastest measured route to the mid target on every dataset
(Table~\ref{tab:headline}): $34\%$ sooner to 0.82 on CIFAR-100, $36\%$
sooner to 0.968 on CIFAR-10 (in 4/5 seeds), $10\%$ sooner to 0.69 on Tiny
ImageNet, and its 3-of-5 median to SVHN's 0.98 numerically edges even full
SAM+Muon's all-seed time. If the mid target is all you need, use it, not
Temperon.

\lead{What it cannot do}
It cannot reach the tier the Muon refiner buys: 0.83 on CIFAR-100 in no
seed (Temperon: 3/5, at 4870s), 0.97 on CIFAR-10 in no seed (Temperon: 2/5),
targets no SGD-refined method touches at all; and its SVHN final sits
0.09pp below Temperon ($p{=}0.025$).

\lead{And on Tiny ImageNet it wins outright}
It matches Temperon's final accuracy (0.7020 vs 0.7003, $p{=}0.42$) and
reaches every late target sooner ($-13/-10/-7\%$ at 0.68/0.69/0.70). The
reason is structural. Tiny ImageNet is the one dataset where the Muon
refiner does not help (full SAM+Muon lands 1.8pp \emph{below} both, at
0.6838), so Temperon's tail there is plain SAM+SGD and there is no tier
for it to retreat to. Where the expensive refiner buys nothing, the simpler
allocation is the better method; Temperon keeps the early game (0.65 at
1322s vs 4533s, the explorer's head start) and nothing at the top. This is
the measured boundary of the method, stated as part of the result:
\textbf{Temperon's win condition is a task where the expensive refiner buys
a tier.}

\begin{takeaway}{Rival takeaway}
Late-phase SAM is the right tool below the SGD-refined ceiling on every
dataset; Temperon exists for the tier above it. Where no such tier exists
(Tiny ImageNet), the rival wins, and we say so.
\end{takeaway}

\section{Transfer: GPT-2 pretraining}\label{sec:lm}

\begin{table}[t]
  \centering\small
  \caption{\textbf{GPT-2 124M, WikiText-103, 400M tokens} (one seed,
  back-to-back runs on the same GPU). The tail arm \emph{is} the late-phase
  allocation run under Muon; there is no separate rival row
  (\S\ref{sec:rival}).}
  \label{tab:lm}
  \begin{tabular}{lccc}
    \toprule
    arm & val.\ loss & ppl & wall-clock \\
    \midrule
    Muon, no SAM & 3.3182 & 27.61 & \textbf{3079s} \\
    SAM+Muon, full-time & 3.3101 & 27.39 & 5335s \\
    \rowcolor{hlrowblue}
    SAM tail \emph{(ours)} & \textbf{3.3051} & \textbf{27.25} & 3796s $\cdot$ $-29\%$ \\
    \bottomrule
  \end{tabular}
\end{table}

\begin{figure}[t]
  \centering
  \includegraphics[width=0.72\linewidth]{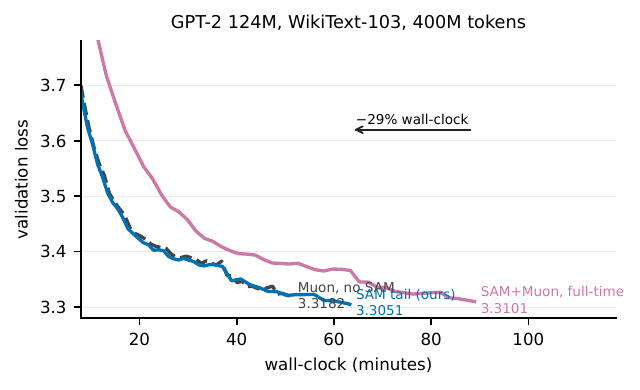}
  \caption{\textbf{The allocation law on GPT-2.} Validation loss against
  wall-clock; the tail matches full-time SAM (gap inside the measured noise
  floor) at $-29\%$, and beats it by 0.061 nats at equal wall-clock.}
  \label{fig:lm}
\end{figure}

On this task the two methods of \S\ref{sec:rival} coincide. Transformers
offer no viable cheap/expensive \emph{optimizer} pair: plain SGD does not
train them, and every adaptive step costs roughly the same. The only
expensive mode left is SAM's second pass, and Temperon reduces to the
late-phase allocation run under Muon: one WSD schedule, SAM switched on at
the decay boundary (the last 30\% of steps), $\rho$ ramped over 400 steps.
The tail matches full-time SAM at $-29\%$ wall-clock and beats it by 0.063
nats at equal wall-clock (Figure~\ref{fig:lm}); the 0.005-nat final gap is
inside the noise floor, which we measure for free by exploiting that the
no-SAM and tail arms are bit-identical before the switch. What this section
adds to \citet{zhou2025latephase} is transfer: the allocation law survives
the move to LM pretraining, to a Muon base, and to a 30\% budget against
vision's 57\%.

\lead{Is SAM worth it here at all}
Marginally: full-time SAM buys 0.008 nats over plain Muon for $+73\%$
wall-clock. And not under data repetition: forcing 20 passes over a
20M-token slice makes SAM \emph{worse} than no-SAM (5.196 vs 5.014 val
loss) while the allocation equivalence still holds (tail 5.198 vs full
5.196). The law is about \emph{where} SAM spends, independently of whether
SAM pays.

\section{Transfer: GLUE fine-tuning}\label{sec:glue}

\begin{table}[t]
  \centering\small
  \caption{\textbf{GLUE fine-tuning} (RoBERTa-base, 5 seeds, best dev
  score, $\rho{=}0.05$). Measured cost ratios: tail $1.28\times$ and full
  $1.97\times$ the no-SAM run, against the designed $1.30/2.00$; tail vs
  full is $-36\%$ wall-clock.}
  \label{tab:glue}
  \begin{tabular}{lcccc}
    \toprule
    task & no SAM & SAM tail & SAM full-time & tail $-$ full \\
    \midrule
    RTE   & 0.7819 & \textbf{0.7870} & 0.7596 & $+2.74$pp, $p{=}0.047$ \\
    MRPC  & 0.9097 & \textbf{0.9158} & 0.9026 & $+1.31$pp, $p{=}0.002$ \\
    STS-B & \textbf{0.9090} & 0.9089 & 0.9045 & $+0.44$pp, $p{=}0.004$ \\
    CoLA  & 0.6204 & \textbf{0.6248} & 0.6133 & $+1.14$pp, $p{=}0.219$ \\
    \midrule
    avg   & 0.8052 & \textbf{0.8091} & 0.7950 & \\
    \bottomrule
  \end{tabular}
\end{table}

\begin{table}[t]
  \centering\small
  \caption{\textbf{The $\rho$ correction.} Tail-vs-full contrasts before and
  after correcting the perturbation radius. RTE's margin, flagged fragile
  in advance by its noise floor (0.0181, dev $n{=}277$), disappears; MRPC
  survives.}
  \label{tab:rho}
  \begin{tabular}{lcc}
    \toprule
    tail $-$ full & at $\rho{=}0.05$ & at $\rho{=}0.02$ \\
    \midrule
    MRPC  & $+1.31$pp ($p{=}0.002$) & $+1.11$pp ($p{=}0.023$) \\
    STS-B & $+0.44$pp ($p{=}0.004$) & $+0.24$pp ($p{=}0.063$) \\
    \rowcolor{hlrow}
    RTE   & $+2.74$pp ($p{=}0.047$) & $-0.43$pp ($p{=}0.516$) \\
    \bottomrule
  \end{tabular}
\end{table}

At face value the tail does not just match full-time SAM at $-36\%$ cost
--- it beats it on all four tasks (Table~\ref{tab:glue}). We do not make
that claim, for two measured reasons. First, full-time SAM at $\rho{=}0.05$
is \emph{harmful} here, worse than no SAM on all four tasks, so part
of the tail's margin is ``harms less'', not ``helps more''; this fails to
reproduce the positive fine-tuning results of \citet{bahri2022sam} at our
scale and budget, and a $\rho$ sweep ($0.05/0.02/0.01$) shows smaller radii
stop the damage but never turn it into a gain. Second, re-running the tail
at the corrected $\rho{=}0.02$ shrinks its edge on every task and removes
it on the noisiest (Table~\ref{tab:rho}). The claim that holds at both
radii is \textbf{equivalence at a third of the SAM cost}: the tail is never
worse than full-time SAM on any task at either $\rho$, paying SAM on 30\%
of steps: the same shape as pretraining, in a regime where SAM itself
does not pay.

\begin{takeaway}{Fine-tuning takeaway}
At a corrected perturbation radius the tail's accuracy edge survives only
on MRPC; what holds everywhere is equivalence at a third of the SAM cost.
We report the weaker, true claim.
\end{takeaway}

\section{Why the switch is scheduled}\label{sec:scheduled}

Could the hand-off be adaptive instead? We tested the two candidate
decisions offline, over the logged trajectories of this paper's runs, with
rules, parameter grids and read-out criteria pre-registered before any
replay ran.

\lead{Timing}
A dimensionless stopping rule --- switch when the explorer's trailing
improvement rate falls below a fraction $c$ of its own historical average
--- was replayed over the pure-cheap trajectories on a locked grid. On
vision it fires at epochs 16--35 against a known-good 43, and the few grid
cells that clear the pre-registered accuracy constraints all fail
cross-seed stability (firing spreads up to $\pm33$ epochs). The failure is
structural, not parametric: under a cosine schedule the accuracy curve is
plateau-then-surge (Figure~\ref{fig:plateau} shows it even for full-time
SAM+Muon), so any backward-looking rule reads the mid-run plateau as
exhaustion and stops before a surge it cannot see. The information that
places the switch lives in the schedule's future, not the trajectory's
past, which is also why a switch inside the decay loses and the
anneal-aligned schedule wins. The one place the rule works is WSD, whose
stable phase genuinely has diminishing returns: there it fires at
0.61--0.70 of the budget against the manual 0.70.

\lead{Refiner choice}
Can early observables predict whether Muon will buy a tier? Not within a
run: the observable signals overlap across datasets, and the strongest of
them \emph{inverts}: early Muon dominance over SGD is largest on exactly
the dataset where SAM+Muon later loses. The information exists one level
up: Muon's hand-off plateau as a fraction of the cheap optimizer's
\emph{final} accuracy separates Tiny ImageNet cleanly (0.90--0.92 against
$\ge$0.958 elsewhere, threshold $\approx$0.94); but the denominator is
precisely the quantity the timing result says cannot be forecast. As a
two-run protocol (a baseline run plus a 43\%-budget probe) the refiner
choice is practical; as a single-run online decision it is not.

\begin{takeaway}{Scheduling takeaway}
The information that places the switch lives in the schedule's future, not
the trajectory's past; the scheduled anneal-aligned hand-off is not a
simplification of an adaptive method --- it is the method.
\end{takeaway}

\section{Honest scope}\label{sec:scope}

Measured limits, stated because they define where the method applies.

\begin{itemize}[leftmargin=1.2em, itemsep=2pt]
\item \textbf{The cheap optimizer wins below its own ceiling.} Plain SGD
  reaches 0.80 on CIFAR-100 in 1567s; Temperon needs 4416s. Temperon never
  accelerates a target a cheap method can already reach.
\item \textbf{A cheaper allocation wins below \emph{its} ceiling too.}
  Late-phase SAM reaches the mid target sooner on every dataset; the win
  starts above the SGD-refined ceiling (0.8234 on CIFAR-100, 0.9688 on
  CIFAR-10), which no arm without a Muon refiner crosses.
\item \textbf{Where the refiner buys nothing, the rival wins outright}
  (Tiny ImageNet, \S\ref{sec:rival}).
\item \textbf{Saturated tasks gain nothing} (SVHN,
  Figure~\ref{fig:plateau}): when the target sits below the expensive
  method's mid-run plateau, the anneal (and therefore the allocation)
  has nothing to sell.
\item \textbf{Data scarcity is not task difficulty.} Twenty passes over a
  20M-token slice made SAM worse, not better (\S\ref{sec:lm}).
\item \textbf{SAM does not help RoBERTa fine-tuning at any $\rho$ we
  tested} (\S\ref{sec:glue}); the fine-tuning claim is about allocation,
  not about SAM helping.
\item \textbf{Scale.} All vision results are one 27.6M CNN at 100 epochs;
  the LM is a 124M single-seed pilot with a measured noise floor. The
  transfer evidence we have is diversity of regime (CNN pretraining, LM
  pretraining, transformer fine-tuning), not scale.
\end{itemize}

\section{Related work}\label{sec:related}

\lead{Sharpness and its cost}
SAM \citep{foret2021sam} and its adaptive variant \citep{kwon2021asam}
double per-step cost; \citet{bahri2022sam} report fine-tuning gains for
language models, which we fail to reproduce at our scale
(\S\ref{sec:glue}). \citet{zhou2025latephase} is the closest work and the
strongest rival: they establish that SAM's benefit concentrates late in
training; we re-run their method at matched budget on four datasets
(\S\ref{sec:rival}), extend the allocation law to a Muon base, to LM
pretraining and to fine-tuning, and report the dataset where their method
beats ours.

\lead{Optimizer switching}
SWATS \citep{keskar2017swats} switches Adam to SGD on a convergence trigger
and \emph{estimates the incoming learning rate from the outgoing
trajectory}: prior art for both the hand-off skeleton and its re-entry
machinery, though it allocates no expensive mode. Warm restarts
\citep{loshchilov2017sgdr} supply the cyclic explorer we ultimately found
unnecessary; stochastic weight averaging \citep{izmailov2018swa} is another
member of the pay-late family. Our own earlier line arrived here by
elimination: a game-theoretic PPO selector choosing among same-cost
optimizers per epoch \citep{mastromichalakis_players} measured no better
than random switching, and its successor OptiRoulette
\citep{mastromichalakis_optiroulette} adopted random switching as the
mechanism, over a per-optimizer-type learning-rate table, with the
switch-failure mitigations (scaled momentum transfer, grace periods) that
this paper's re-entry ramps descend from. Collapsing that decision space to
one scheduled switch is where the present work begins.

\lead{Muon, and ranking under partial budgets}
Muon \citep{jordan2024muon} was scaled to LLM training by
\citet{liu2025muon}. Concurrent work \citep{zhong2026spectral} also pairs
SAM with Muon, at ImageNet scale, matching the perturbation's geometry to
Muon's spectral update: it asks which geometry the ascent step should live
in, while we ask where in the run it should be paid for. The two questions
compose --- our allocation law is agnostic to the perturbation geometry the
tail uses --- and their result independently confirms the pairing's
strength beyond the scales measured here. Finally, the inversion of
\S\ref{sec:scheduled} --- the
expensive method looking \emph{best} early exactly where it fails late ---
is a systematic instance of the crossing-curves failure mode that
successive-halving searchers \citep{li2018hyperband} assume away; ranking
anneal-paying methods at partial budget will systematically prune the
eventual winners.

\section{Conclusion}\label{sec:conclusion}

Temperon is not a new optimizer; it is a measured answer to where expensive
training modes belong: in a tail that owns the entire final anneal, handed
to the strongest refiner the task rewards. Applied as stated, the recipe
reaches full-time-SAM quality at roughly a third less wall-clock across
three regimes (vision pretraining, GPT-2 pretraining at $-29\%$, GLUE
fine-tuning at $-36\%$), and its economics are predictable before training
starts: skipping SAM early buys a fixed credit, a Muon tail pays a flat
$1.50\times$ premium against a SAM+SGD epoch, and the practitioner chooses
once whether the credit returns as time or as accuracy.

The claims are deliberately narrower than the results table might tempt.
The accuracy contribution is the Muon refiner, not our allocation shape or
its restarts, and we withdrew both as contributions when the ablations said
so. The closest rival is faster to every mid-level target and wins Tiny
ImageNet outright; Temperon's win condition is a task where the expensive
refiner buys a tier the cheaper methods cannot reach. And the schedule is
not a placeholder for a smarter adaptive rule: under cosine schedules the
trajectory carries no stopping signal, so the anneal-aligned scheduled
switch is, on present evidence, the method rather than an approximation of
one. We consider these boundary measurements part of the deliverable; each
one tells a practitioner when \emph{not} to use the recipe, which is
information a benchmark table alone does not carry.

What remains open is scale and generality. Everything here lives on a 27.6M
CNN, a 124M language model and RoBERTa-base; whether the allocation law
survives ImageNet-class vision and larger language models is the obvious
next measurement. The credit/premium framework itself never mentions SAM,
which suggests testing it on other expensive modes, such as weight
averaging or resolution schedules. And the two-run refiner test of
\S\ref{sec:scheduled} is one out-of-sample validation away from becoming an
automatic mode. The recipe ships today as a pip-installable wrapper around
any pair of \texttt{torch} optimizers
(\texttt{pip install temperon}), together with the full experimental
record, including the negative controls this paper rests on.

\bibliographystyle{plainnat}
\bibliography{refs}

\end{document}